\documentclass[runningheads]{llncs}
\usepackage{graphicx}

\usepackage{tikz}
\usepackage{comment}
\usepackage{amsmath,amssymb} 
\usepackage{color}

\usepackage[accsupp]{axessibility}  

\usepackage{booktabs}
\usepackage{multirow}
\usepackage[normalem]{ulem}
\useunder{\uline}{\ul}{}
\usepackage{hyperref}

\begin{document}
\pagestyle{headings}
\mainmatter
\def\ECCVSubNumber{13}  

\title{Frequency Dropout: Feature-Level Regularization via Randomized Filtering} 

\titlerunning{Frequency Dropout}
%

\author{Mobarakol Islam \and
Ben Glocker}
%
\authorrunning{M. Islam and B. Glocker}
%
\institute{BioMedIA Group, Department of Computing, Imperial College London, UK 
\email{(m.islam20, b.glocker)@imperial.ac.uk}\\}
\maketitle

\begin{abstract}
Deep convolutional neural networks have shown remarkable performance on various computer vision tasks, and yet, they are susceptible to picking up spurious correlations from the training signal. So called `shortcuts' can occur during learning, for example, when there are specific frequencies present in the image data that correlate with the output predictions. Both high and low frequencies can be characteristic of the underlying noise distribution caused by the image acquisition rather than in relation to the task-relevant information about the image content. Models that learn features related to this characteristic noise will not generalize well to new data.

In this work, we propose a simple yet effective training strategy, Frequency Dropout, to prevent convolutional neural networks from learning frequency-specific imaging features. We employ randomized filtering of feature maps during training which acts as a feature-level regularization. In this study, we consider common image processing filters such as Gaussian smoothing, Laplacian of Gaussian, and Gabor filtering. Our training strategy is model-agnostic and can be used for any computer vision task. We demonstrate the effectiveness of Frequency Dropout on a range of popular architectures and multiple tasks including image classification, domain adaptation, and semantic segmentation using both computer vision and medical imaging datasets. Our results suggest that the proposed approach does not only improve predictive accuracy but also improves robustness against domain shift.

\end{abstract}

\section{Introduction}
\label{sec:intro}

The impressive performance of deep convolutional neural networks (CNNs) in computer vision is largely based on their ability to extract complex predictive features from images that correlate well with the prediction targets such as categorical image labels. If the training data, however, contains spurious correlations, for example, between characteristics of image acquisition and image annotations, there is a high risk that the learned features will not generalize to new data acquired under different conditions. A key issue is that features related to such spurious correlations are often much easier to learn and can be trivially picked up via convolution kernels. Specific sensor noise, for example, may manifest itself as high or low frequency patterns in the image signal. When trained on such biased data, CNNs may establish so called `shortcuts' instead of learning generalizable, task-specific feature representations which may be more difficult to extract.

The issue of shortcut learning \cite{geirhos2020shortcut} and related aspects of texture bias in computer vision have been discussed in great detail in previous work~\cite{geirhos2018imagenet,zhang2019identity}. Recently, Wang et al.~\cite{wang2020high} observe that high-frequency feature components cause a trade-off between accuracy and robustness. Low-level image characteristics are generally easier to pick up and thus lead to a much quicker decrease in the loss function early during training \cite{achille2018critical}.

Several theoretical and empirical studies tried to tackle this issue by synthesizing shape-based representation of the dataset~\cite{geirhos2018imagenet}, informative dropout~\cite{shi2020informative}, pooling geometry~\cite{cohen2016inductive}, smoothing kernels~\cite{wang2020high}, or antialiasing via two step pooling~\cite{azulay2018deep}. Curriculum by smoothing (CBS) \cite{sinha2020curriculum} introduces a curriculum-based feature smoothing approach to reduce the use of high-frequency features during training. The approach controls the high-frequency information propagated through a CNN by applying a Gaussian filter on the feature maps during training. The curriculum consists of decreasing the standard deviation of the Gaussian filter as training progresses. While this avoids the use of high frequency features at the beginning of the training procedure, the CNN may still pick up these features at a later stage.

In this paper, we propose a simple yet effective training strategy, Frequency Dropout (FD), preventing CNNs from learning frequency-specific imaging features by employing randomized feature map filtering. We utilize three different types of filters including Gaussian smoothing, Laplacian of Gaussian, and Gabor filters with randomized parameters. Similar to dropout, these filters are applied randomly during training but act as feature-level regularization instead of dropping activations. FD can be incorporated into any architecture and used across a wide range of applications including image classification and semantic segmentation.

Our main findings are as follows:
\begin{itemize}
  \item The proposed Frequency Dropout yields consistent improvements in predictive accuracy across a range of popular CNN architectures;
  \item Besides improved accuracy, we also observe improved robustness for networks trained with FD when tested on corrupted, out-of-distribution data;
  \item FD improves results in various tasks including image classification, domain adaptation, and semantic segmentation demonstrating its potential value across a wide range of computer vision and medical imaging applications;
\end{itemize}

Frequency Dropout can be easily implemented in all popular deep learning frameworks. It is complementary to other techniques that aim to improve the robustness of CNNs such as dropout, data augmentation, and robust learning.

\section{Related work}

\subsection{Image filtering in CNNs}

Incorporating image filtering such as smoothing, sharpening, or denoising within the training of CNNs has been widely explored and shown potential for improving model robustness and generalization. The main use of image filtering is for input-level data augmentation~\cite{khosla2020supervised,taori2020measuring,hossain2021robust,lopes2019improving,dai2020contrastively} and feature-level normalization~\cite{sinha2020curriculum,zhang2019making,lee2020compounding,mairal2016end,azulay2019deep}. SimCLR~\cite{chen2020simple} uses filters such as Gaussian noise, Gaussian blur, and Sobel filters to augment the images for contrastive learning. Taori et al.~\cite{taori2020measuring} study model robustness under controlled perturbations using various types of filters to simulate distribution shift. Laplacian networks~\cite{lassance2021laplacian} uses Laplacian smoothing to improve model robustness.
Gabor filters have been considered within CNNs to encourage orientation- and scale-invariant feature learning~\cite{alekseev2019gabornet,luan2018gabor,perez2020gabor}. Gaussian blurring is also utilized in self-supervised learning~\cite{navarro2021evaluating}, and domain adaptation~\cite{dai2020contrastively}. In terms of CNN feature normalization, smoothing filters are used as anti-aliased max-pooling~\cite{zhang2019making}, and anti-alias downsampling~\cite{lee2020compounding,mairal2016end} to improve the internal feature representation. Most recently, anti-aliasing filtering has been used to smooth the CNN feature maps in a curriculum learning manner with a reported increase in performance in various vision tasks~\cite{sinha2020curriculum}.

\subsection{Dropout-based regularization}

Dropout~\cite{srivastava2014dropout,hinton2012improving} has been widely used as regularization technique to prevent overfitting~\cite{wager2013dropout,srivastava2014dropout}, pruning~\cite{gomez2018targeted,salehinejad2019ising}, spectral transformation~\cite{khan2019regularization} and uncertainty estimation~\cite{gal2016dropout,nair2020exploring}. Monte Carlo dropout~\cite{gal2016dropout} is used to estimate prediction uncertainty at test time. Targeted dropout~\cite{gomez2018targeted} omits the less useful neurons adaptively for network pruning. Dropout has also been explored for data augmentation by projecting dropout noise into the input space~\cite{bouthillier2015dropout}. Spatial dropout~\cite{amini2018spatial} proposes 2D dropout to knock out full kernels instead of individual neurons in convolutional layers.

\section{Background and preliminaries}
\label{background}

In CNNs, a convolution layer is used to extract features from input data $x$ by convolving ($\circledast$) the input with a kernel $w$. A convolution layer in a typical CNN may be followed by a pooling and activation layer (e.g., a ReLU) which can be expressed as

\begin{equation}
    out := ReLU(pool(w \circledast x))
\end{equation}

\begin{figure*}[!htbp]
    \centering
    \includegraphics[width=0.98\textwidth]{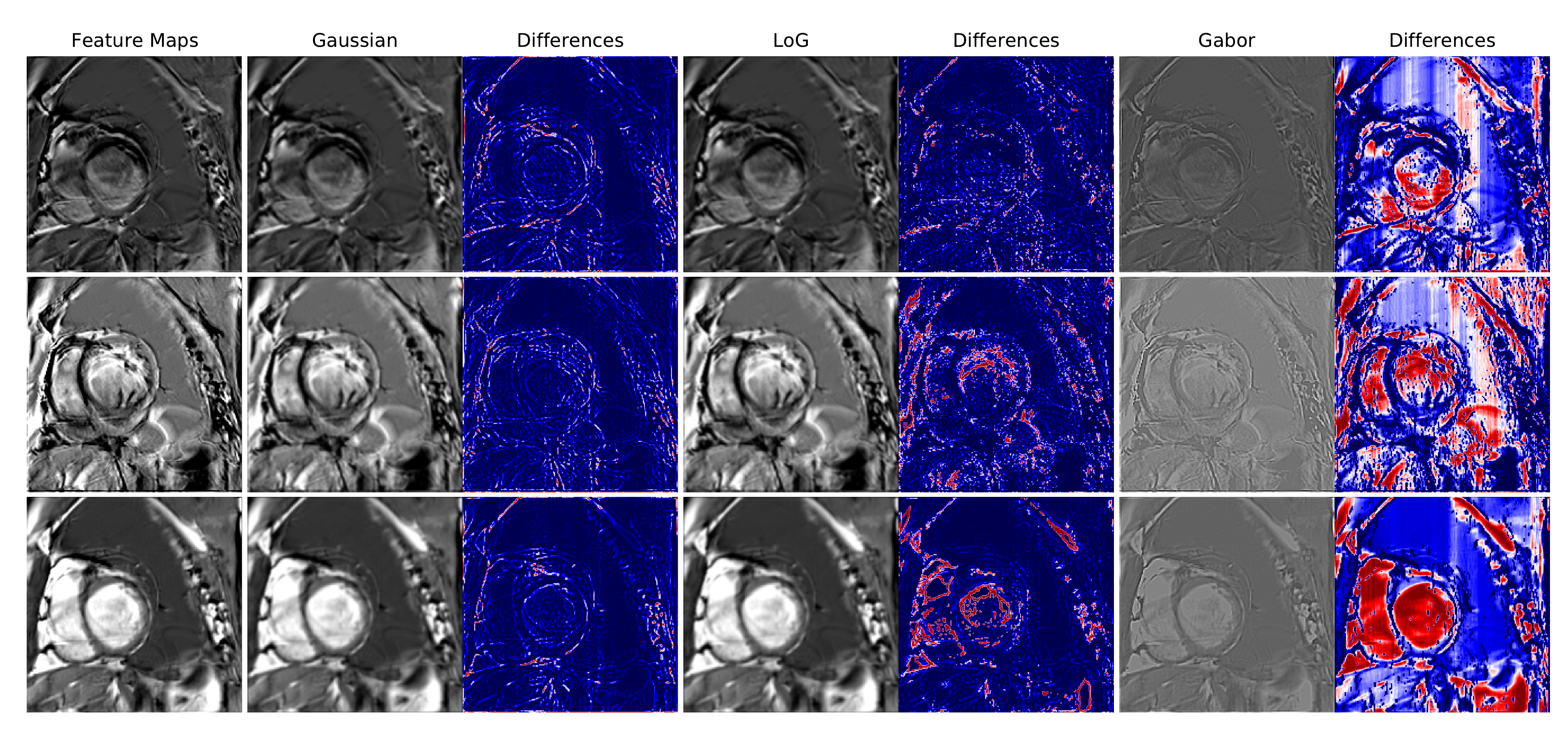}
    \caption{Regularization of feature maps by applying filters such as Gaussian, Laplacian of Gaussian, and Gabor. The differences of feature maps before and after filtering is also illustrated in beside each filter.}
    \label{fig:filtered_feature_maps}
\end{figure*}

\subsection{Image filters}
\label{sec:filters}
Image filters are commonly used in digital image processing, often for smoothing or sharpening of the original image. For example, Gaussian, Laplacian of Gaussian (LoG), Gabor filters are widely used for image smoothing, sharpening, and enhancing. In general, a filter is applied via spatial convolution of the image using a specifically designed filter kernel. We can categorize image filters considered in this work into low-pass and high-pass filters. 

The Gaussian filter has a blurring effect by removing high-frequency components of the image signal. The Gaussian filter is characterized by the kernel size and standard deviation. A 2D Gaussian kernel (G) with standard deviation $\sigma$ can be written as

\begin{equation}
    G(x,y|\sigma) = -{\frac{1}{2\pi \sigma^2}}{e^{-\frac{x^2+y^2}{2\sigma^2}}}
    \label{equ:log}
\end{equation}

Laplacian of Gaussian (LoG) is a hybrid filter with Gaussian smoothing and a Laplacian kernel. It smooths the image and enhances the edges (or regions of strong intensity changes). A zero centered 2D LoG kernel with a standard deviation of $\sigma$ can be expressed as
\begin{equation}
    LoG(x,y|\sigma) = -{\frac{1}{\pi \sigma^4}}{\left[1-\frac{x^2+y^2}{2\sigma^2}\right]}{e^{-\frac{x^2+y^2}{2\sigma^2}}}
    \label{equ:log}
\end{equation}

The Gabor filter is a special case of a band filter where a Gaussian kernel is modulated by a sinusoidal signal of a particular frequency and orientation. It is used in the application of edge detection, texture analysis, and feature extraction in both spatial and frequency domains. There are several parameters such as wavelength ($\lambda$) and phase ($\psi$) of the sinusoidal function, standard deviation ($\sigma$) of Gaussian kernel, and spatial aspect ratio ($\gamma$) of the function, controlled the effect of the filter. The real component of the Gabor kernel (Ga) can be represented as below.
\begin{equation}
Ga(x,y|\sigma)=\exp \left(-\frac{x^{\prime 2}+\gamma^{2} y^{\prime 2}}{2 \sigma^{2}}\right) \cos \left(2 \pi \frac{x^{\prime}}{\lambda}+\psi\right)
\end{equation}

\subsection{Spatial dropout}
Dropout~\cite{srivastava2014dropout,hinton2012improving} refers to randomly dropping out neurons during the training phase. The probability of dropout can be controlled with a parameter $p$. Based on $p$, dropout generates either 0 or 1 sampled from a Bernoulli distribution to keep or drop the neurons using simple multiplication. If dropout of $p$ probability is applied on a layer ($l$) of input  $y^{(l)}$ then dropout can be formulated as below.

\begin{flalign}
r^{(l)} &= Bernoulli(p)\\
\hat{y}^{(l)} &= r^{(l)} \ast y^{(l)}
\end{flalign}

\section{Frequency Dropout}

We design a feature-level regularization technique by randomizing the choice of filter and its parameters. We utilize three filter types including Gaussian, Laplacian of Gaussian, and Gabor filter within a feature regularization layer that can be incorporated into any CNN architecture. More specifically, a randomized filter from the regularization layer is applied with a certain probability after each convolution operation to post-process the generated feature maps.

A filter is selected randomly in each training iteration and each CNN layer to suppress different frequencies. The randomized filter selection can be formulated as below.
\begin{flalign}
\label{equ:filter}
RF &:= \textrm{rand.choice}[G,\ LoG,\ Ga]
\end{flalign}


\subsection{FD with randomized filtering}

Given a randomly selected filter, we also randomly sample filter parameters from specific ranges which affect different frequencies in the input signal. We use dropout to turn off kernels at random locations, adopting the strategy of spatial dropout \cite{amini2018spatial,tompson2015efficient} to turn off the entire kernel instead of individual neurons. The dropout probability can vary for each filter type ($p^G,\ p^{LoG},\ p^{Gabor}$ ). When a kernel is turned off it means the input feature map in this position will remain unchanged after the frequency dropout layer. If  $\sigma^{(n)}$ is the vector of randomly generated frequencies (or standard deviation) to obtain kernels of $n$ channels using a selected filter ($randomized\_filter$ from equation \ref{equ:filter}), then Frequency Dropout by Randomized Filtering (FD-RF) can be formulated as follows

\begin{flalign}
\sigma^{(n)} &:= \text{rand}(n, \ low, \ high)\\
\sigma_{fd}^{(n)} &:= \text{dropout}(\sigma^{(n)}, \ p)\\
w_{fd}^{(n)} &:= RF(\sigma_{fd}^{(n)})
\end{flalign}

where $low$ and $high$ are the limits of the random frequencies for a filter and $p$ is the dropout probability which can be pre-defined as $p^G,\ p^{LoG},\ p^{Gabor}$ for Gaussian, LoG and Gabor, respectively. Other notations can be denoted as random frequency $\sigma^{(n)}$, spatial frequency dropout $\sigma_{fd}^{(n)}$ and kernel with frequency dropout $w_{fd}^{(n)}$. Fig.~\ref{fig:filtered_feature_maps} illustrates the effect of these filters on feature maps.

An simple example of the layer-wise application of Frequency Dropout with Randomized Filtering (FD-RF) for a few convolutional layers on input data $x$ and convolution kernel $w$ is as follows:
\begin{flalign}
w_{fd(i)}^{(n)} &:= RF(\cdot|\sigma_{fd(i)}^{(n)}) \nonumber\\
layer_i &:=ReLU(pool(w_{fd(i)}^{(n)} \circledast (w \circledast x_i))) \nonumber \\
w_{fd(i+1)}^{(n)} &:= RF(\cdot|\sigma_{fd(i+1)}^{(n)}) \nonumber\\
layer_{i+1} &:=ReLU(pool(w_{fd}^{(n)} \circledast (w \circledast x_{i+1}))) \nonumber\\
w_{fd(i+2)}^{(n)} &:= RF(\cdot|\sigma_{fd(i+2)}^{(n)}) \nonumber\\
layer_{i+2} &:=ReLU(pool(w_{fd}^{(n)} \circledast (w \circledast x_{i+2}))) \nonumber
\end{flalign}

\subsection{FD with Gaussian filtering}
To investigate the effectiveness of Frequency Dropout over CBS~\cite{sinha2020curriculum}, we additionally define a simplified version of FD-RF with a fixed Gaussian filter. As CBS uses Gaussian filtering in a curriculum manner, Frequency Dropout by Gaussian filtering (FD-GF) correspondingly applies feature map smoothing during training but in a randomized fashion rather than following a curriculum strategy. The example layer-wise application from above would simply change accordingly to: 
\begin{flalign}
w_{fd(i)}^{(n)} &:= G(\cdot|\sigma_{fd(i)}^{(n)}) \nonumber\\
layer_i &:=ReLU(pool(w_{fd(i)}^{(n)} \circledast (w \circledast x_i))) \nonumber\\
w_{fd(i+1)}^{(n)} &:= G(\cdot|\sigma_{fd(i+1)}^{(n)}) \nonumber\\
layer_{i+1} &:=ReLU(pool(w_{fd}^{(n)} \circledast (w \circledast x_{i+1}))) \nonumber\\
w_{fd(i+2)}^{(n)} &:= G(\cdot|\sigma_{fd(i+2)}^{(n)}) \nonumber\\
layer_{i+2} &:=ReLU(pool(w_{fd}^{(n)} \circledast (w \circledast x_{i+2}))) \nonumber
\end{flalign}

\section{Experiments and results}
\label{experiments}

We conduct extensive validation of our proposed Frequency Dropout technique including the tasks of image classification and semantic segmentation in the settings of supervised learning and unsupervised domain adaptation. We utilize several state-of-the-art neural network architectures for computer vision and medical imaging applications to demonstrate the effectiveness of our approach. All quantitative metrics are produced from running experiments with two different random seeds. The average is reported as the final metric.

\subsection{Image classification}

Due to its simplicity and flexibility, we could easily integrate FD into various state-of-the-art classification networks including ResNet-18~\cite{he2016deep}, Wide-ResNet-52~\cite{zagoruyko2016wide}, ResNeXt-50~\cite{xie2017aggregated} and VGG-16~\cite{simonyan2014very}. We conduct all experiments on three classification datasets including CIFAR-10, CIFAR-100~\cite{krizhevsky2009learning} and SVHN~\cite{goodfellow2013multi}. To measure the robustness of our method, we test the performance of the trained classification models on CIFAR-10-C and CIFAR-100-C which are corrupted dataset variations~\cite{hendrycks2019benchmarking}.

\begin{table*}[!h]
\centering
\caption{Classification accuracy of our FD over CBS~\cite{sinha2020curriculum} and Baseline. Boldface indicates the top two models with higher performance and additional underline for the best model. All the experiments are conducted on a common dropout probability of $p^G=0.4,\ p^{LoG}=0.5,\ p^{Gabor}=0.8$}. 
\scalebox{0.98}{
\begin{tabular}{c|c|c|c|c|l|l}
\hline
\multicolumn{1}{l|}{}            & \multicolumn{1}{l|}{}         & \multicolumn{3}{c|}{Classification}                                                     & \multicolumn{2}{c}{Robustness}                                   \\ \hline
\multicolumn{1}{l|}{}            & \multicolumn{1}{l|}{}         & CIFAR-100                   & CIFAR-10                    & SVHN                        & \multicolumn{1}{c|}{CIFAR-100-C} & \multicolumn{1}{c}{CIFAR-10-C} \\ \hline
                                 & Baseline                      & 65.31 ± 0.14                & 89.24 ± 0.23                & 96.26 ± 0.06                & 43.68 ± 0.17                     & 70.36 ± 0.75                   \\ \cline{2-7} 
                                 & CBS                           & 65.77 ± 0.45                & 89.81 ± 0.09                & 96.27 ± 0.06                & 46.69 ± 0.04                     & 74.17 ± 0.29                   \\ \cline{2-7} 
                                 & FD-GF                         & \textbf{67.45 ± 0.54}       & \textbf{90.33 ± 0.47}       & {\ul \textbf{96.70 ± 0.30}} & \textbf{46.78 ± 0.30}            & \textbf{74.32 ± 3.15}          \\ \cline{2-7} 
\multirow{-4}{*}{\rotatebox{90}{ResNet-18}}      & FD-RF & {\ul \textbf{68.20 ± 0.52}} & {\ul \textbf{90.53 ± 0.28}} & \textbf{96.60 ± 0.03}       & {\ul \textbf{48.10 ± 0.07}}      & {\ul \textbf{74.43 ± 1.31}}    \\ \hline
                                 & Baseline                      & 58.54 ± 0.35                & 87.53 ± 0.08                & 95.77 ± 0.05                & 40.10 ± 0.01                     & 71.58 ± 0.33                   \\ \cline{2-7} 
                                 & CBS                           & \textbf{63.67 ± 0.11}       & \textbf{89.47 ± 0.14}       & {\ul \textbf{96.39 ± 0.02}} & \textbf{44.53 ± 1.33}            & {\ul \textbf{74.41 ± 0.45}}    \\ \cline{2-7} 
                                 & FD-GF                         & 63.09 ± 0.14                & 89.32 ± 0.27                & 96.26 ± 0.13                & {\ul \textbf{44.69 ± 1.22}}      & \textbf{74.01 ± 1.20}          \\ \cline{2-7} 
\multirow{-4}{*}{\rotatebox{90}{VGG-16}}         & FD-RF & {\ul \textbf{63.94 ± 0.35}} & {\ul \textbf{89.59 ± 0.07}} & \textbf{96.26 ± 0.01}       & 44.26 ± 0.29                     & 73.12 ± 1.97                   \\ \hline
                                 & Baseline                      & 68.06 ± 0.07                & {\ul \textbf{90.98 ± 0.24}} & 97.04 ± 0.05                & 39.63 ± 0.52                     & 67.39 ± 0.28                   \\ \cline{2-7} 
                                 & CBS                           & 65.04 ± 0.45                & 87.00 ± 1.53                & {\ul \textbf{97.12 ± 0.02}} & 37.05 ± 4.22                     & 65.45 ± 1.64                   \\ \cline{2-7} 
                                 & FD-GF                         & {\ul \textbf{68.38 ± 0.59}} & 90.80 ± 0.45                & 97.05 ± 0.11                & \textbf{41.00 ± 0.18}            & \textbf{69.22 ± 1.17}          \\ \cline{2-7} 
\multirow{-4}{*}{\rotatebox{90}{W-ResNet}} & FD-RF & \textbf{68.11 ± 0.32}       & \textbf{90.88 ± 0.12}       & \textbf{97.09 ± 0.16}       & {\ul \textbf{41.39 ± 0.66}}      & {\ul \textbf{69.63 ± 0.30}}    \\ \hline
                                 & Baseline                      & \textbf{68.88 ± 0.59}       & {\ul \textbf{90.84 ± 0.30}} & 96.31 ± 0.02                & 44.72 ± 0.47                     & 68.85 ± 0.86                   \\ \cline{2-7} 
                                 & CBS                           & 65.15 ± 1.27                & 89.28 ± 0.40                & 96.15 ± 0.12                & 42.46 ± 1.67                     & 69.68 ± 0.37                   \\ \cline{2-7} 
                                 & FD-GF                         & 68.35 ± 0.45                & 89.69 ± 0.44                & {\ul \textbf{96.63 ± 0.01}} & \textbf{45.12 ± 1.73}            & {\ul \textbf{72.17 ± 1.60}}    \\ \cline{2-7} 
\multirow{-4}{*}{\rotatebox{90}{ResNeXt}}     & FD-RF & {\ul \textbf{68.95 ± 0.66}} & \textbf{89.86 ± 0.47}       & \textbf{96.47 ± 0.05}       & {\ul \textbf{46.24 ± 0.56}}      & \textbf{70.36 ± 0.68}          \\ \hline
\end{tabular}}
\label{tab:classification}
\end{table*}

\begin{figure}[!htb]
    \centering
    \includegraphics[width=0.68\textwidth]{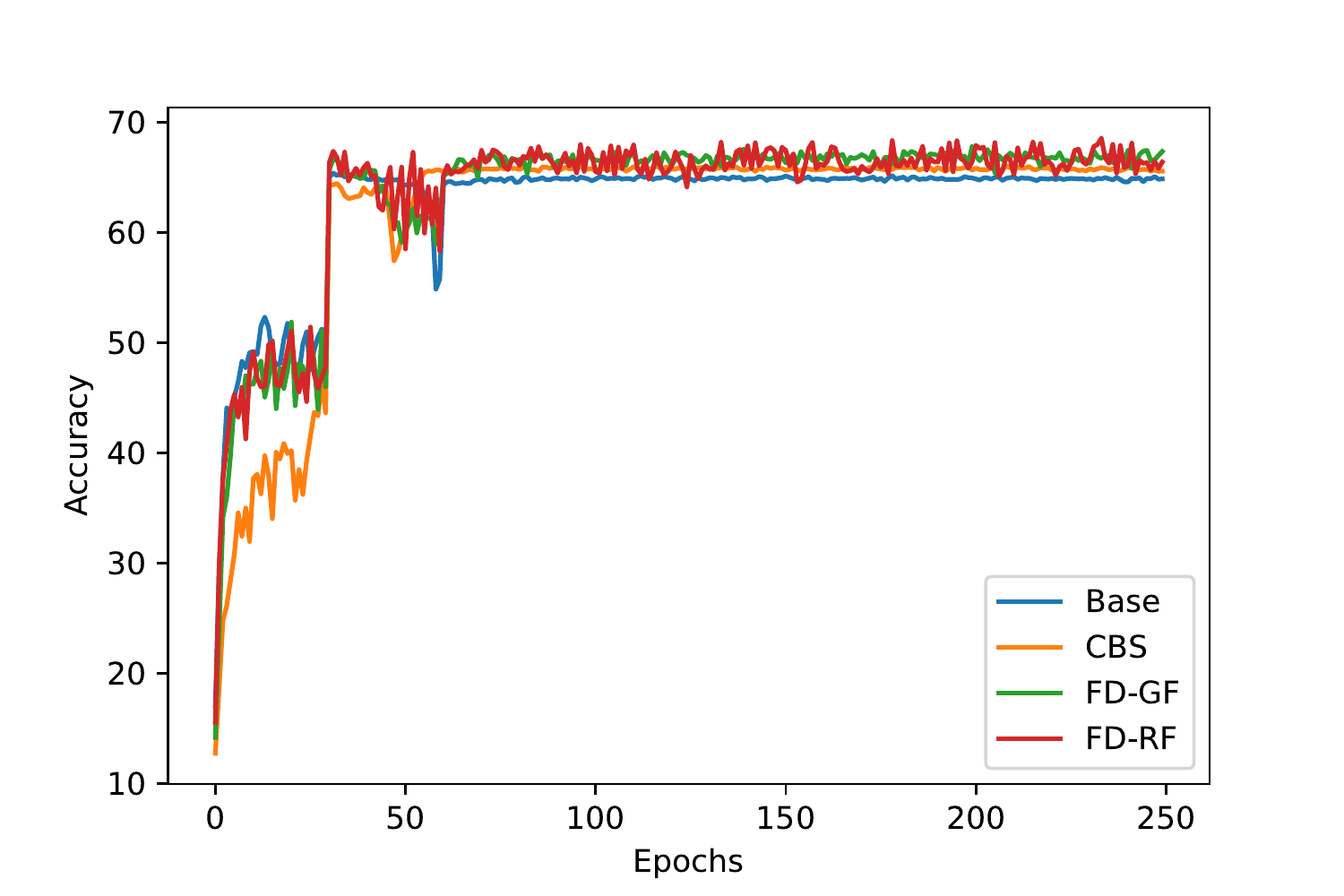}
    \caption{Validation performance over the epochs for ResNet-18 with CIFAR-100.}
    \label{fig:vailation_acc}
\end{figure}

\begin{figure*}[!htbp]
    \centering
    \includegraphics[width=0.98\textwidth]{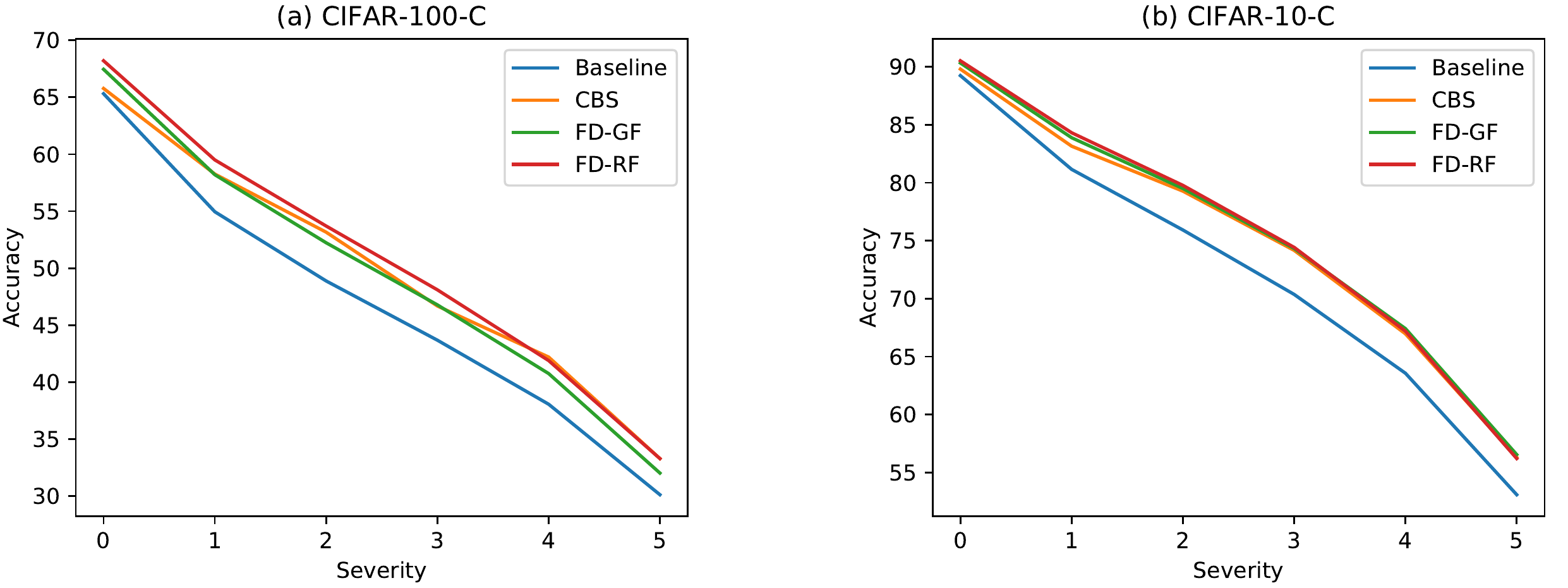}
    \caption{Robustness plots with 0 to 5 severity for (a) CIFAR-100-C and (b) CIFAR-10-C. The result obtains on ResNet-18 for Baseline, CBS, FD-GF, and FD-RF. The proposed method shows constant performance preservation over baseline where CBS preserves robustness with higher severity.}
    \label{fig:cifarc_severity}
\end{figure*}

The original implementation\footnote{\url{https://github.com/pairlab/CBS}} of the closely related technique of Curriculum by Smoothing (CBS)~\cite{sinha2020curriculum} is adopted for comparison with FD and the baseline models. In CBS, all classification models are optimized with SGD and the same settings for learning rate schedule, weight decay, and momentum throughout all experiments. We use a common set of dropout parameter ($p^G=0.4,\ p^{LoG}=0.5,\ p^{Gabor}=0.8$) for all the experiments which is tuned with ResNet-18 and CIFAR-100 dataset. Fig.~\ref{fig:vailation_acc} shows the validation accuracy versus epochs for ResNet-18 on CIFAR-100. Table~\ref{tab:classification} presents the classification and robustness performance metrics of our method compared to baselines and CBS. There is a significant improvement in classification accuracy for FD-RF and FD-GF over the baseline and CBS for most of the architectures and datasets. Specifically, FD improves 2-3\% accuracy for ResNet-18 with the CIFAR-100 classification dataset. The table also shows the superior performance of our method on corrupted CIFAR-10-C and CIFAR-100-C datasets for the median severity. Robustness performance against severity level is shown in Fig.~\ref{fig:cifarc_severity}. Both FD-RF and FD-GF show robust prediction accuracy for different severity levels over baseline, while CBS does show competitive performance in higher severity levels of corruption.

\subsection{Domain adaptation}

\subsubsection{Vision application}

We investigate the performance of FD with the task of unsupervised domain adaptation (UDA) in visual recognition. We use three datasets of MNIST~\cite{lecun1998mnist}, USPS~\cite{hull1994database} and SVHN~\cite{goodfellow2013multi} with four architectures of ResNet-18~\cite{he2016deep}, Wide-ResNet-52~\cite{zagoruyko2016wide}, ResNeXt-50~\cite{xie2017aggregated} and VGG-16~\cite{simonyan2014very} by following domain setup of~\cite{tzeng2017adversarial} where a model train on one dataset and test on different data source for the task of UDA in classification. The datasets consist of 10 classes and arrange into three directions of adaptation. The UDA performance of FD, CBS, and baselines are shown in Table \ref{tab:uda_classification}. There is 3-4\% performance improvement with our method for multiple architectures such as ResNet-18, VGG-16, Wide-ResNet-52 for the adaptation settings of MNIST→USPS, USPS→MNIST, and SVHN→MNIST, respectively. Overall, FD-GF and FD-RF obtain the best performance compared to baseline and CBS for most of the cases.

\begin{table*}[!h]
\centering
\caption{Unsupervised domain adaptation (UDA) with classification task. Boldface indicates the top two models with higher accuracy and additional underline for the best model.}
\scalebox{0.98}{
\begin{tabular}{c|c|c|c|c}
\hline
\multicolumn{1}{l|}{}           & \multicolumn{1}{l|}{} & MNIST-\textgreater{}USPS    & USPS-\textgreater{}MNIST    & SVHN-\textgreater{}MNIST    \\ \hline
\multirow{4}{*}{ResNet-18}      & Baseline              & 82.06 ± 0.27                & {\ul \textbf{77.18 ± 0.08}} & 79.88 ± 0.20                \\ \cline{2-5} 
                                & CBS                   & 84.30 ± 0.46                & 50.82 ± 0.33                & 82.20 ± 0.25                \\ \cline{2-5} 
                                & FD-GF                 & \textbf{85.35 ± 0.27}       & 66.48 ± 0.23                & {\ul \textbf{82.66 ± 0.21}} \\ \cline{2-5} 
                                & FD-RF                 & {\ul \textbf{86.60 ± 0.20}} & \textbf{73.76 ± 0.25}       & \textbf{82.43 ± 0.19}       \\ \hline
\multirow{4}{*}{VGG-16}         & Baseline              & {\ul \textbf{84.20 ± 0.20}} & 52.03 ± 0.33                & 78.68 ± 0.36                \\ \cline{2-5} 
                                & CBS                   & 78.95 ± 0.12                & 49.61 ± 0.26                & {\ul \textbf{83.35 ± 0.20}} \\ \cline{2-5} 
                                & FD-GF                 & 82.81 ± 0.35                & \textbf{54.07 ± 0.11}       & 78.67 ± 0.25                \\ \cline{2-5} 
                                & FD-RF                 & \textbf{84.11 ± 0.14}       & {\ul \textbf{57.67 ± 0.19}} & \textbf{79.22 ± 0.14}       \\ \hline
\multirow{4}{*}{Wide-ResNet-52} & Baseline              & 79.32 ± 0.36                & \textbf{86.81 ± 0.22}       & \textbf{83.97 ± 0.50}       \\ \cline{2-5} 
                                & CBS                   & 75.99 ± 0.20                & 85.52 ± 0.26                & 82.26 ± 0.17                \\ \cline{2-5} 
                                & FD-GF                 & \textbf{82.96 ± 0.12}       & 86.64 ± 0.07                & 81.71 ± 0.19                \\ \cline{2-5} 
                                & FD-RF                 & 79.17 ± 0.23                & {\ul \textbf{88.53 ± 0.11}} & {\ul \textbf{86.02 ± 0.28}} \\ \hline
\multirow{4}{*}{ResNeXt-50}     & Baseline              & \textbf{92.62 ± 0.18}       & {\ul \textbf{60.04 ± 0.14}} & {\ul \textbf{83.20 ± 0.38}} \\ \cline{2-5} 
                                & CBS                   & 88.23 ± 0.25                & 51.83 ± 0.13                & 81.97 ± 0.18                \\ \cline{2-5} 
                                & FD-GF                 & {\ul \textbf{92.87 ± 0.08}} & \textbf{59.98 ± 0.11}       & \textbf{83.12 ± 0.27}       \\ \cline{2-5} 
                                & FD-RF                 & 91.53 ± 0.19                & 52.12 ± 0.19                & 80.83 ± 0.22                \\ \hline
\end{tabular}}
\label{tab:uda_classification}
\end{table*}


\subsubsection{Medical application}

To evaluate the performance of FD with a real-world medical dataset, we utilize the Multi-Centre, Multi-Vendor, and Multi-Disease Cardiac Segmentation (M\&MS) MRI dataset~\cite{campello2021multi}. It contains 150 cases each from two vendors of A and B equally. The annotation consists of three cardiac regions including left ventricle (LV), right ventricle (RV), and myocardium (MYO). We split the data vendor-wise for train and validation so that the experiments reflect the domain shift setup. A 3D UNet~\cite{cciccek20163d} implementation\footnote{\url{https://github.com/lescientifik/open\_brats2020}} is adopted as the baseline architecture. We integrate our FD-RF, FD-GF, and closely related work CBS~\cite{sinha2020curriculum}. The filtering effects for all these techniques are only introduced in the encoder part as it plays the role of extracting image features while the decoder operates in the space of segmentation. During training, the Adam optimizer is used with a learning rate of $10^{-4}$ and cross-entropy loss. We use cross-validation by swapping vendor-A and vendor-B as train and validation sets. 

\begin{table*}[!h]
\centering
\caption{Cross-validation performance of cardiac image segmentation from 3D MRI. All the experiments conduct on a common dropout probability of $p^G=0.5,\ p^{LoG}=0.5,\ p^{Ga}=0.8$. Cross-vendor validation is done to produce the prediction for both vendors A and B. Boldface indicates the top two models with higher DSC and additional underline for the best model.}
\begin{tabular}{c|c|c|c|c||c|c|c|c}
\hline
\multicolumn{1}{l|}{} & \multicolumn{4}{c||}{Vendor-A  to Vendor-B}                                                & \multicolumn{4}{c}{Vendor-B to Vendor-A}                                                 \\ \hline
\multicolumn{1}{l|}{} & LV                   & RV                   & MYO                  & \textbf{Mean DSC}        & LV                   & RV                   & MYO                  & \textbf{Mean DSC}        \\ \hline
Baseline              & 72.19                & \textbf{65.43}       & 60.28                & 65.97                & 52.98                & 39.30                & \textbf{42.08}       & 44.79                \\ \hline
CBS                   & 71.60                & 62.28                & 62.24                & 65.37                & \textbf{63.11}       & {\ul \textbf{46.40}} & 31.33                & 46.95                \\ \hline
FD-GF                 & {\ul \textbf{72.47}} & 63.05                & \textbf{64.01}       & \textbf{66.51}       & {\ul \textbf{63.26}} & 44.34                & 40.45                & \textbf{49.35}       \\ \hline
FD-RF                 & \textbf{68.96}       & {\ul \textbf{66.30}} & {\ul \textbf{67.71}} & {\ul \textbf{67.66}} & 51.47                & \textbf{44.57}       & {\ul \textbf{47.56}} & {\ul \textbf{51.47}} \\ \hline
\end{tabular}
\label{tab:uda_seg_cv}
\end{table*}

\begin{figure*}[!htb]
    \centering
    \includegraphics[width=0.98\textwidth]{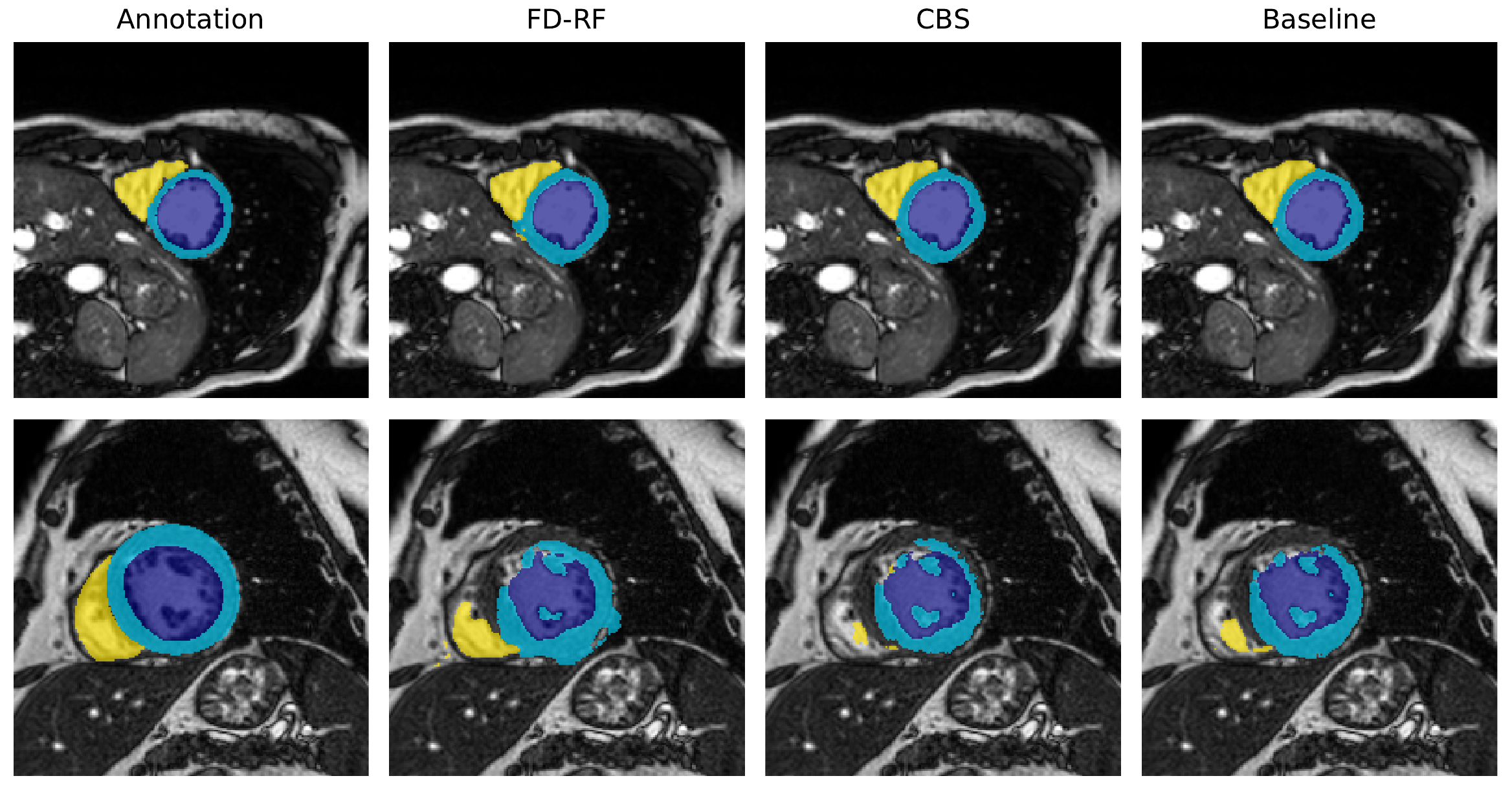}
    \caption{UDA performance in segmentation for Baseline, CBS, and FD-RF. The middle slices of two random MRI scans are visualized for annotation and prediction on different models. The colors of purple, blue and yellow indicate the left ventricle (LV), myocardium (MYO), and right ventricle (RV).}
    \label{fig:segmentation_visualization}
\end{figure*}

The results of cross-vendors validation are reported in the Table \ref{tab:uda_seg_cv}. Dice similarity coefficient (DSC) is used to measure the segmentation prediction for baseline, CBS, and our FD-GF, FD-RF. The left side of the table contains model performances trained on vendor-A and validation on vendor-B and swapping the vendors on the right side. The results suggest a 2-3\% increase of the mean DSC for our method over CBS where 4-6\% improvement compared to baseline. The prediction visualization is also showing better segmentation for FD-RF compared to other methods in Figure~\ref{fig:segmentation_visualization}. Overall, the results indicate the better generalization and robustness capacity of the proposed method on dataset shift and domain shift.


\subsection{Ablation Study}

\subsubsection{Dropout Ratio}
\begin{figure}[!h]
    \centering
    \includegraphics[width=0.7\textwidth]{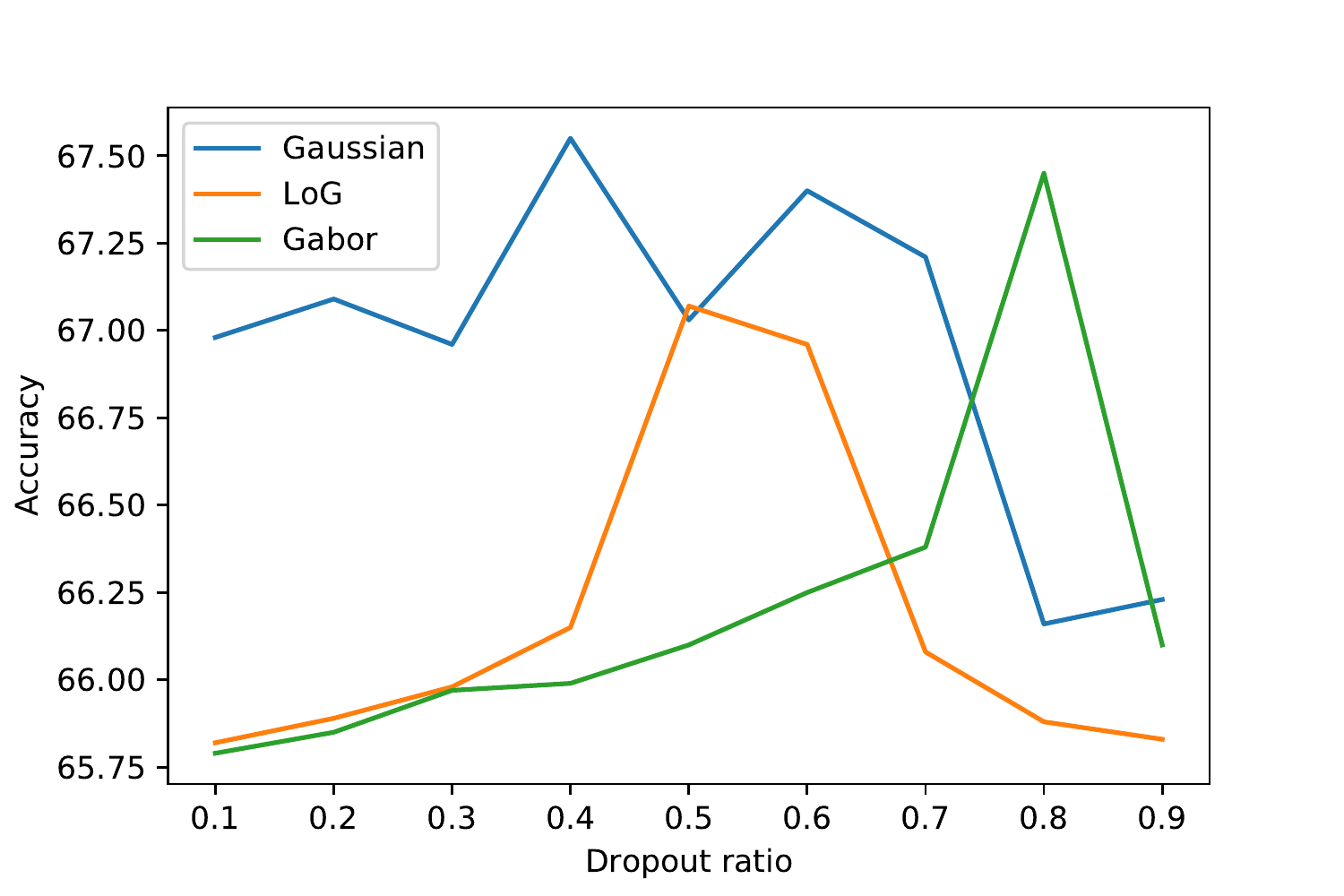}
    \caption{Effect of individual filtering and selecting dropout ratio for our FD technique. The optimal dropout ratios for each filter obtain as $p^G=0.4,\ p^{LoG}=0.5,\ p^{Ga}=0.8$ for ResNet-18 with CIFAR-100.}
    \label{fig:dropout_ratio}
\end{figure}

To determine the effective dropout ratio for each filter, we perform specific experiments by varying the ratio for the individual filter. Fig.~\ref{fig:dropout_ratio} shows the accuracy over dropout ratio in the range 0.1 to 0.9 for ResNet-18 with CIFAR-100. From this, we selected dropout ratios of Gaussian ($p^G=0.4$), LoG ($p^{LoG}=0.5$), and Gabor ($p^{Ga}=0.8$) filters for all further experiments of classification tasks.

\subsubsection{Kernel size}

The considered imaging filters can be used with different kernel sizes. We investigate the FD performance under four different kernel sizes including [$1\times1,\ 3\times3,\ 5\times5,\ 7\times7$]. Table~\ref{tab:kernel_size} presents the performance of FD by varying kernel size. Kernel size of $3\times3$ seems most effective in the design of FD for regularizing feature maps and is used throughout all other experiments.

\begin{table}[!h]
\centering
\caption{Effect of different kernel size for FD using ResNet-18 and CIFAR dataset.}
\begin{tabular}{c|c|c|c|c}
\hline
\multicolumn{1}{l|}{} & 1x1   & 3x3   & 5x5   & 7x7   \\ \hline
CIFAR-100             & 65.34 & \textbf{68.57} & 68.00 & 67.19 \\ \hline
CIFAR-10              & 89.15 & \textbf{90.33} & 89.94 & 89.88 \\ \hline
\end{tabular}
\label{tab:kernel_size}
\end{table}

\section{Discussion and conclusion}

We introduced a novel method for feature-level regularization of convolutional neural networks namely Frequency Dropout with Randomized Filtering (FD-RF). We considered Gaussian, Laplacian of Gaussian, and Gabor filters. We provide empirical evidence across a larger number of experiments showing a consistent improvement of CNN performance on image classification, semantic segmentation, and unsupervised domain adaptation tasks on both computer vision and a real-world medical imaging dataset. We also observe improvements for model robustness on the corrupted CIFAR datasets (CIAFR-10-C and CIFAR-100-C). To make a fair comparison with closely related work of CBS~\cite{sinha2020curriculum}, we build a simplified version of FD-RF with only Gaussian filters (FD-GF). Our experimental results suggest that FD yields significant improvements over baselines and CBS in most of the cases for both computer vision and medical datasets. In terms of robustness with CIAFR-10-C and CIFAR-100-C, FD-RF notably outperforms the baseline for all levels of severity. CBS shows competitive accuracy in higher severity compared to FD especially for CIFAR-10-C (see in Fig.~\ref{fig:cifarc_severity}). We also observe that due to feature-level regularization, FD needs longer training where baseline and CBS converge after fewer epoch (see in Fig.~\ref{fig:vailation_acc}). The future direction of this work is to consider additional image filters. It may also be beneficial to consider different types of filters at different depths of the CNN. For example, some types of filters may be more effective in earlier layers of the network whereas some filters may be harmful when applied to the feature maps in later layers. An interesting direction would be to combine the randomized filtering with a curriculum learning approach where either the probability of selecting certain filters or their parameter ranges are annealed as training progresses.

\paragraph{\textbf{Acknowledgements.}}
This project has received funding from the European Research Council (ERC under the European Union’s Horizon 2020 research and innovation programme (Grant Agreement No. 757173, Project MIRA).

%
%
\bibliographystyle{splncs04}
\bibliography{references}
\end{document}